\pdfoutput=1

\documentclass{article}
\usepackage{ijcai24}

\usepackage{times}
\usepackage{soul}
\usepackage{url}
\usepackage[hidelinks]{hyperref}
\usepackage[utf8]{inputenc}
\usepackage[small]{caption}
\usepackage{graphicx}
\graphicspath{ {./images/} }
\usepackage{amsmath}
\usepackage{amsthm}
\usepackage{booktabs}
\usepackage{algorithm}
\usepackage{algorithmic}
\usepackage[switch]{lineno}

\title{GigaPevt: Multimodal Medical Assistant}

\author{
Pavel Blinov$^1$
\and
Konstantin Egorov$^1$\and
Ivan Sviridov$^1$\and
Nikolay Ivanov$^1$\and
Stepan Botman$^1$\and
Evgeniy Tagin$^1$\and
Stepan Kudin$^1$\and
Galina Zubkova$^1$\And
Andrey Savchenko$^1$\\
\affiliations
$^1$Sber AI Lab\\
\emails
\{Blinov.P.D, Egorov.K.Ser, IAnatoSviridov, NikIvanov, SABotman, EyTagin, SSKudin, GVZubkova, AVladSavchenko\}@sber.ru,
}

\begin{document}

\maketitle

\begin{abstract}
    Building an intelligent and efficient medical assistant is still a challenging AI problem. The major limitation comes from the data modality scarceness, which reduces comprehensive patient perception. This demo paper presents the GigaPevt, the first multimodal medical assistant that combines the dialog capabilities of large language models with specialized medical models. Such an approach shows immediate advantages in dialog quality and metric performance, with a 1.18\% accuracy improvement in the question-answering task.
\end{abstract}

\section{Introduction}
The availability of high-quality healthcare remains a significant social problem for many countries due to high-cost issues, shortage of medical staff, or inaccessibility in far-end regions. From a technological point of view, one possible solution lies in developing intelligent medical assistants. There are many attempts to build such an assistant in the form of a symptom checker application (Ada Health, K Health, Symptomate, Babylon Health)~\cite{wallace2022diagnostic}. The major limitation of such systems is using the only text modality, where patients describe the symptoms and get answers in the chat. The system misses a significant part of the non-verbal doctor-patient communication context in that case. Another flaw of such AI assistants is the design in a survey systems paradigm with predefined questions that miss the human touch in the communication process.

However, the rapid development of large language models (LLMs) opens new ways to create next-generation assistants in healthcare. In this research, we introduce an innovative multimodal medical assistant called GigaPevt (Figure~\ref{GigaPevtUI}).\footnote{The {\it Giga}Pevt name is a wordplay in many facets. First, the {\it GP} stands for General Practitioner. Second, in Russian, the word consonant to a therapist with the smaller metric prefix {\it Giga-} instead of {\it Tera-}. Third, it references the core Russian LLM {\it Giga}Chat.} The GigaPevt combines specialized medical models with the rich dialog capabilities of LLM, which account for broad context in visual, audio, and text modalities. That allows us to mitigate the problems mentioned above, significantly improves patient interaction experience, and boosts the quality of dialog.

\begin{figure}
  \includegraphics[width=0.48\textwidth]{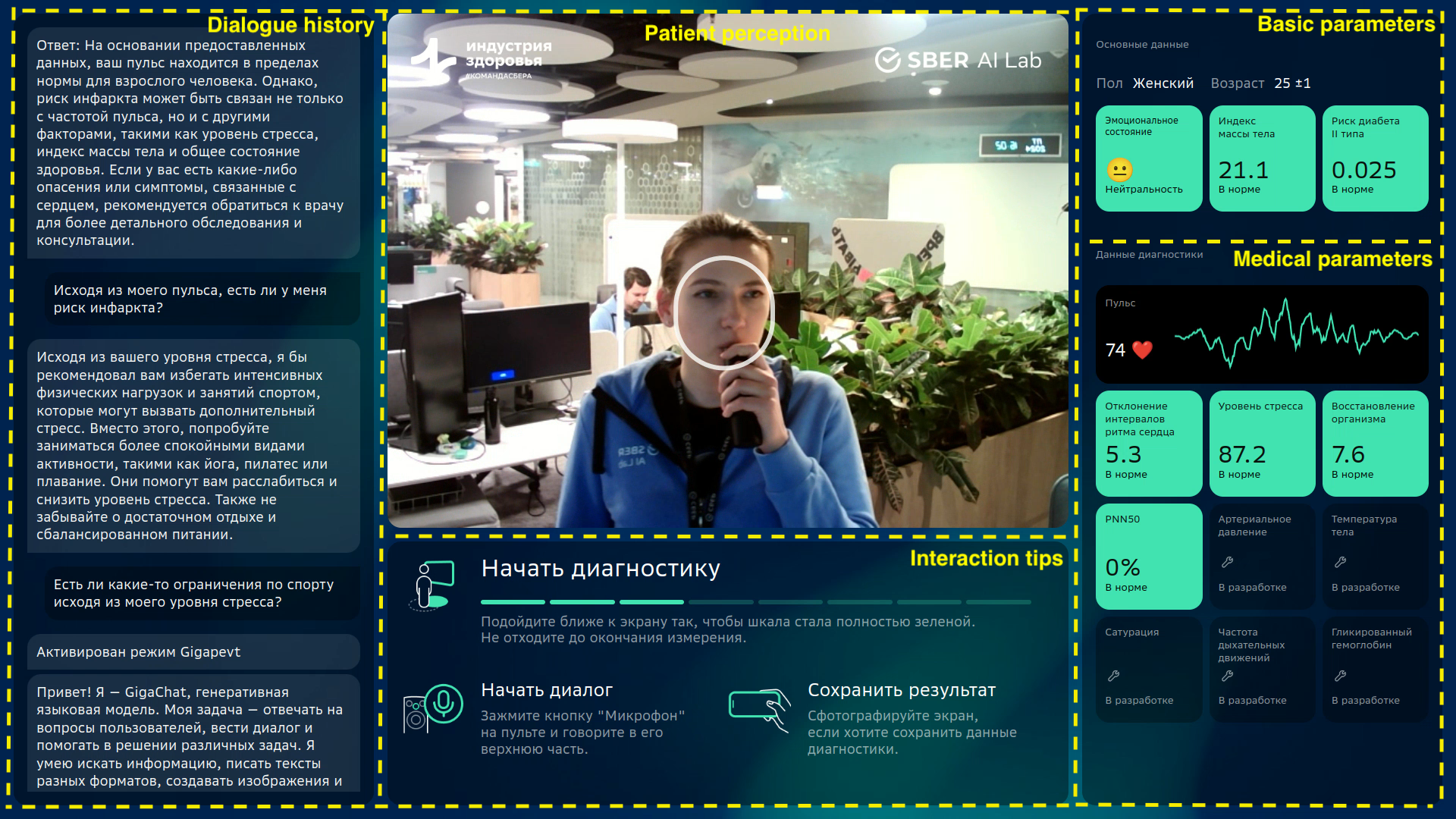}
  \caption{GigaPevt UI}
\end{figure} \label{GigaPevtUI}

The GigaPevt demonstration video is available at \url{https://youtu.be/yETMJqIThPE}.

\section{GigaPevt architecture}
Given the assortment of potential deployment scenarios, we implemented the GigaPevt as a client-server application, as shown in Figure~\ref{architecture}. We use a rich client written in Python to run lightweight models that require low latency for a comfortable user experience. These are the face detector, Text-to-Speech (TTS), and Automatic Speech Recognition (ASR) API services.\footnote{\url{https://developers.sber.ru/docs/ru/salutespeech/category-overview}} Client logic handles data flow, client state management, and network interactions with the rest of the system.

We use Flask as a back-end framework that handles the server side. The server has a model manager, specialized models, and the GP Dialog Logic component responsible for the patient-assistant interaction.







\begin{figure}
  \includegraphics[width=0.48\textwidth]{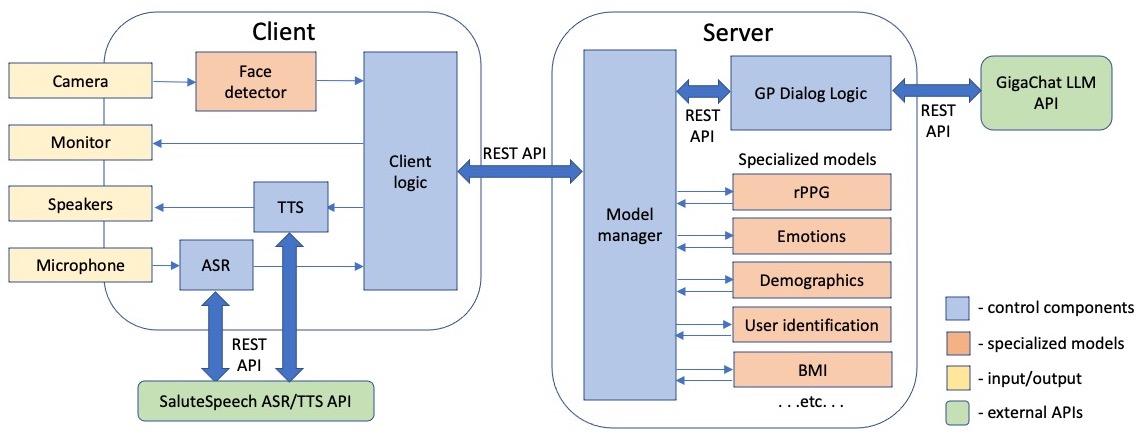}
  \caption{GigaPevt architecture}
\end{figure} \label{architecture} 

\section{Specialized models}
\subsection{Video-based Facial Analytics}
We analyze facial videos in a frame-wise way to identify a user and predict the attributes list. The facial region is detected using the MediaPipe\footnote{\url{https://developers.google.com/mediapipe}} library, and the obtained facial image is fed into the corresponding neural network.

\subsubsection{User Identification}
We apply open-source face recognition model to identify, authorize, and manage the user session. The facial image is resized to 224x224 and fed into the fast OFAMobileNetV3 model~\cite{10168107} to extract facial descriptors. The latter model was pre-trained on the part of the VGGFace2 dataset~\cite{Cao18}. The training dataset contains feature vectors of known subjects obtained by the same procedure given their facial photos. Extracted facial embeddings are normalized using $L_2$ norm and matched with a training dataset by the k-NN with the Euclidean distance. In the experimental study, we gathered a testing dataset with 426 photos of 104 persons. As a result, the recognition accuracy equals 98.19\%.


\begin{table*}
    \centering
    \begin{tabular}{p{0.07\linewidth} p{0.27\linewidth} p{0.27\linewidth} p{0.27\linewidth}}
        \toprule
        Question & GigaChat answer & LLaMA answer & GigaPevt answer \\
        \midrule
        Is my heart rate comfortable for exercise? & To determine a comfortable heart rate when exercising, you need to consider your age and level of physical fitness. Generally, for people aged 20 to 59 years, exercising at a heart rate between 50\% and 85\% of your maximum heart rate is recommended. & The normal heart rate for an adult is between 60 and 100 beats per minute. If your heart rate is within this range, has a regular rhythm, and you do not feel discomfort during exercise, it may be comfortable for you. However, consult your doctor for testing and advice if you have any doubts. & No, your heart rate is currently not comfortable for exercise. Your heart rate is 115 beats per minute, which is higher than normal. For comfortable physical exercise, it is recommended to have a heart rate ranging from 60 to 100 beats per minute. I suggest you consult a doctor to determine the optimal level of physical activity for you. \\
        \midrule
        What do you think? Am I obese? & To determine if you are obese, you need to know your weight and height. Obesity is defined as excess accumulation of fat, which can lead to various diseases and poor quality of life. & As an artificial intelligence model, I cannot determine your weight or estimate your body. However, if you suspect you are obese, it is best to see your doctor to determine your body mass index (BMI) and get advice. The most important thing is not to put off visiting your doctor if you have concerns about your health. & Based on your body mass index, which is within the normal range, you are not obese. However, if you have other signs or symptoms, such as fainting, dizziness, or shortness of breath, it is recommended that you consult your doctor. \\
        \bottomrule
    \end{tabular}
    \caption{Examples of answers to health-related questions (translated from Russian)}
    \label{tab:dialog}
\end{table*}

\subsubsection{Socio-Demographic Model}
We used a high-performance MobileNet-V1 model to predict age, gender, and ethnicity~\cite{savchenko2021facial}. It was pre-trained to recognize faces from the VGGFace2 dataset~\cite{Cao18}. Unfortunately, the age groups in this dataset are very imbalanced, so the trained models work incorrectly for the faces of very young or old people. Hence, we decided to add all (15K) images from the Adience dataset~\cite{eidinger2014age}. As the latter contains only age intervals, e.g., (60-100), we put all images from this interval to the average age, i.e., 80. The ethnicity classifier was trained on the subset of the UTKFace dataset~\cite{zhang2017age} with different class weights to achieve better performance for imbalanced classes. As a result, we obtained gender classification accuracy (93.79\%), which is 2.7\% greater than the best-known accuracy of a much larger VGG-16 neural network in the DEX method~\cite{rothe2015dex}. The age prediction MAE (Mean Absolute Error) is equal to 5.74 years, which is 0.7 years better when compared to the DEX. Finally, the accuracy of ethnicity prediction is equal to 87.6\%, which is on par with the best-known models~\cite{savchenko2021facial}.

\subsubsection{Facial Expression Recognition}
The EfficientNet-B0 model was used to predict facial emotions~\cite{savchenko2022mt}. In particular, we pre-trained it to identify faces on the VGGFace2 dataset as in the previous subsection. Next, we add three heads for predicting eight basic expressions (Anger, Contempt, Disgust, Fear, Happiness, Neutral, Sadness, and Surprise), valence, and arousal using the training part of the AffectNet dataset~\cite{mollahosseini2017affectnet}. In contrast to typical face recognition training, we use highly cropped faces in both training stages to remove hair, background, etc., and concentrate on facial expressions only. Thus, we obtained competitive results on the validation part of the AffectNet dataset. In particular, our accuracy for eight classes equals 61.93\%, 0.5-2\% greater than that of models of similar size and efficiency~\cite{pourmirzaei2021using,savchenko2021facial}.


\subsubsection{Body Mass Index Model}
The Body Mass Index (BMI) estimation is addressed as an image regression problem. The dataset for this task includes samples (face-BMI value) from Reddit,\footnote{\url{https://www.reddit.com/r/progresspics}} a public social web platform enriched with the FIW-BMI dataset~\cite{JIANG2019183}. The detected faces are resized to $256 \times 256$ and fed into the ResNet34 model~\cite{he2015deep} as a backbone with MSE as a metric and loss function. On the test set, we get the MSE equal to 4 units.

\subsubsection{rPPG Model}
The remote photoplethysmography (rPPG) model allows extracting a photoplethysmography signal using a video of a user’s face. Our system used the Plane Orthogonal-to-Skin (POS) algorithm described in~\cite{7565547}. This algorithm is the strong baseline in the field; it adopts signal processing techniques on subtle changes of the RGB values in the time domain. The algorithm runs based on a face landmark detection system, as it relies on averaging pixel values change over face regions of interest (cheeks, nose, and forehead), which are the most valuable for pulse wave detection~\cite{rppgsurvey}. According to~\cite{liu2023rppgtoolbox}, this method has high accuracy, robustness, and speed. Calculating rPPG also allows us to infer several derived parameters of the state of the cardiovascular system, such as heart rate, RR intervals (intervals between consecutive heartbeats), stress index, PNN50, and many others. All these health indicators allow GigaPevt to determine the user’s condition more accurately and comprehensively, improving the patient interaction experience.

\subsection{GP Dialog Logic}

To manage and control the dialog, we used the GigaChat Pro model\footnote{\url{https://developers.sber.ru/portal/products/gigachat}}. As the assistant continuously receives a stream of additional parameters from sensors, we can substantially enrich the prompts. Even the simple inclusion of such augmented parameters in a text prompt is helpful. But our ultimate goal is to obtain controllable, high-quality dialog focused on the anamnesis collection. To achieve this goal, we resorted to the Retrieval-Augmented Generation (RAG) technique~\cite{NEURIPS2020_6b493230} as it can help supplement the user's request with the found relevant information and the Chain of Thoughts (CoT)~\cite{wei2023chainofthought} as it was successfully used in challenging NLP scenarios~\cite{li2023chain}. We started to assemble a database of medical dialogues and questions. These improvements substantially boost the quality of answers and performance in downstream health-related tasks.

\section{Experiments}
For a diverse assessment of the GigaPevt assistant, we evaluate it from quantitative and qualitative sides.

\subsection{Performance Evaluation}
We evaluate the proposed assistant on the two most relevant downstream tasks from the RuMedBench~\cite{blinov2022rumedbench}.

The RuMedDaNet is a binary question-answering task with 512 test medical-related questions in Russian. As shown in Table 2, the GigaPevt model with the RAG technique substantially outperforms GigaChat and Human assessment by 1.18\% and 0.39\%, respectively. It is the most indicative task for assessing the increase in the quality of dialogue because it is close to our assistant's target scenario.

The RuMedNLI is a task of natural language inference with 1536 Russian text pairs to decide whether the second text in a pair is entailment, contradiction, or neutral to the first. For this task, the performance gain is positive but relatively moderate (0.38\%).

\begin{table}
    \centering
    \begin{tabular}{lrr}
        \toprule
        Model  & RuMedDaNet & RuMedNLI \\
        \midrule
        GigaPevt & {\bf 93.36} & 71.59 \\
        Human & 92.97 & {\bf 85.67} \\
        GigaChat & 92.18 & 71.21 \\
        ChatGPT-3.5 & 89.26 & 61.33 \\
        \bottomrule
    \end{tabular}
    \caption{LLMs performance metrics (Accuracy, \%). The best results are shown in bold.}
    \label{tab:medmench}
\end{table}

\subsection{Dialog Showcases}
Table~\ref{tab:dialog} lists several examples to assess the qualitative gap in the answers between the GigaChat model, open LLaMA~\cite{touvron2023llama} and GigaPevt. The GigaChat and LLaMA answers are the base model answers in a zero-shot mode without additional information. GigaPevt answers is the model with visual input, RAG, and CoT for enriching our prompts. Tuned prompts have the following form: constitution (description of how the model should behave and its role) + visual information about a person (pulse, body mass index, etc.) + postfix (recommendations for model).

We can see that the model with prompt tuning, RAG, visual information from the camera, etc., allows us to answer questions that require a visual inspection immediately. Due to generally more awareness about the patient, our assistant gives more specific and concrete answers.

\section{Conclusion and Future Work}
This demo aims to inspire multimodality research in the medical domain. We demonstrate the prototype of medical assistant technology, which processes a rich multimodal context. Several experiments confirm assistant effectiveness.

We plan to use more advanced knowledge management techniques for LLMs, such as Advanced and Modular RAG~\cite{gao2024retrievalaugmented}, to make our model more complete and focused on the medical domain. Also, our plan includes closer integration with a patient electronic health record for better context awareness and the ability to target more specific questions.

\bibliographystyle{named}
\bibliography{GigaPevt}

\end{document}